\documentclass[letterpaper, 10 pt, journal, twoside]{IEEEtran}
\usepackage{ifpdf}
\usepackage{cite}
\pdfoutput=1
\ifCLASSINFOpdf
 \usepackage[pdftex]{graphicx}
\else
 \usepackage[dvips]{graphicx}
\fi

\usepackage{amsmath}
\usepackage{amsfonts, bm}
\usepackage{algorithmic}
\usepackage{array}
\usepackage{fixltx2e}
\usepackage{stfloats}
\usepackage{url}

\begin{document}
	\title{Robot Cooking with Stir-fry: Bimanual \\
		Non-prehensile Manipulation of Semi-fluid Objects} 
	
	\author{Junjia Liu, Yiting Chen, Zhipeng Dong, Shixiong Wang, \\Sylvain Calinon, Miao Li, and Fei Chen, \IEEEmembership{Senior Member, IEEE}
	
		\thanks{Manuscript received: October, 22, 2021; Revised: December, 13, 2021; Accepted: February, 4, 2022.}
		
		\thanks{This paper was recommended for publication by Editor Markus Vincze upon evaluation of the Associate Editor and Reviewers’ comments. This work was supported in part by the Research Grants Council of the Hong Kong Special Administrative Region, China (Ref. No. 24209021), the VC Fund 4930745 of the CUHK T Stone Robotics Institute and CUHK Direct Grant for Research 4055140. (\textit{Corresponding authors: Miao Li and Fei Chen.)}
		}
		\thanks{Junjia Liu, Zhipeng Dong, Shixiong Wang and Fei Chen are with the Department of Mechanical and Automation Engineering, T-Stone Robotics Institute, The Chinese University of Hong Kong, Hong Kong  (e-mail: jjliu@mae.cuhk.edu.hk; zhipengdongneu@gmail.com; sxwang@hkclr.hk; f.chen@ieee.org).}
		
		\thanks{Yiting Chen is with the School of Power and Mechanical Engineering, Wuhan University, Hubei, China (e-mail: chenyiting@whu.edu.cn).}%
		
		\thanks{Sylvain Calinon is with the Idiap Research Institute, Martigny, Switzerland (e-mail: sylvain.calinon@idiap.ch).}%
		
		\thanks{Miao Li is with the Institute of Technological Sciences, Wuhan University, Hubei, China
				(e-mail: miao.li@whu.edu.cn).}%
		\thanks{Digital Object Identifier (DOI): see top of this page.}
	}

	\markboth{IEEE Robotics and Automation Letters. Preprint Version. Accepted February, 2022}
	{Liu \MakeLowercase{\textit{et al.}}: Robot Cooking with Stir-fry: Bimanual  Non-prehensile Manipulation of Semi-fluid Objects} 
	\maketitle
	\IEEEpeerreviewmaketitle
	
	\begin{abstract}   
		This letter describes an approach to achieve well-known Chinese cooking art stir-fry on a bimanual robot system. Stir-fry requires a sequence of highly dynamic coordinated movements, which is usually difficult to learn for a chef, let alone transfer to robots. In this letter, we define a canonical stir-fry movement, and then propose a decoupled framework for learning this deformable object manipulation from human demonstration. First, dual arms of the robot are decoupled into different roles (a leader and follower) and learned with classical and neural network based methods separately, then the bimanual task is transformed into a coordination problem. To obtain general bimanual coordination, we secondly propose a Graph and Transformer based model --- \textit{Structured-Transformer}, to capture the spatio-temporal relationship between dual-arm movements. Finally, by adding visual feedback of contents deformation, our framework can adjust the movements automatically to achieve the desired stir-fry effect. We verify the framework by a simulator and deploy it on a real bimanual Panda robot system. The experimental results validate our framework can realize the bimanual robot stir-fry motion and have the potential to extend to other deformable objects with bimanual coordination.
	\end{abstract}


\section{Introduction}
\begin{figure}[htp]
	\centering
	\includegraphics[width=0.75\linewidth]{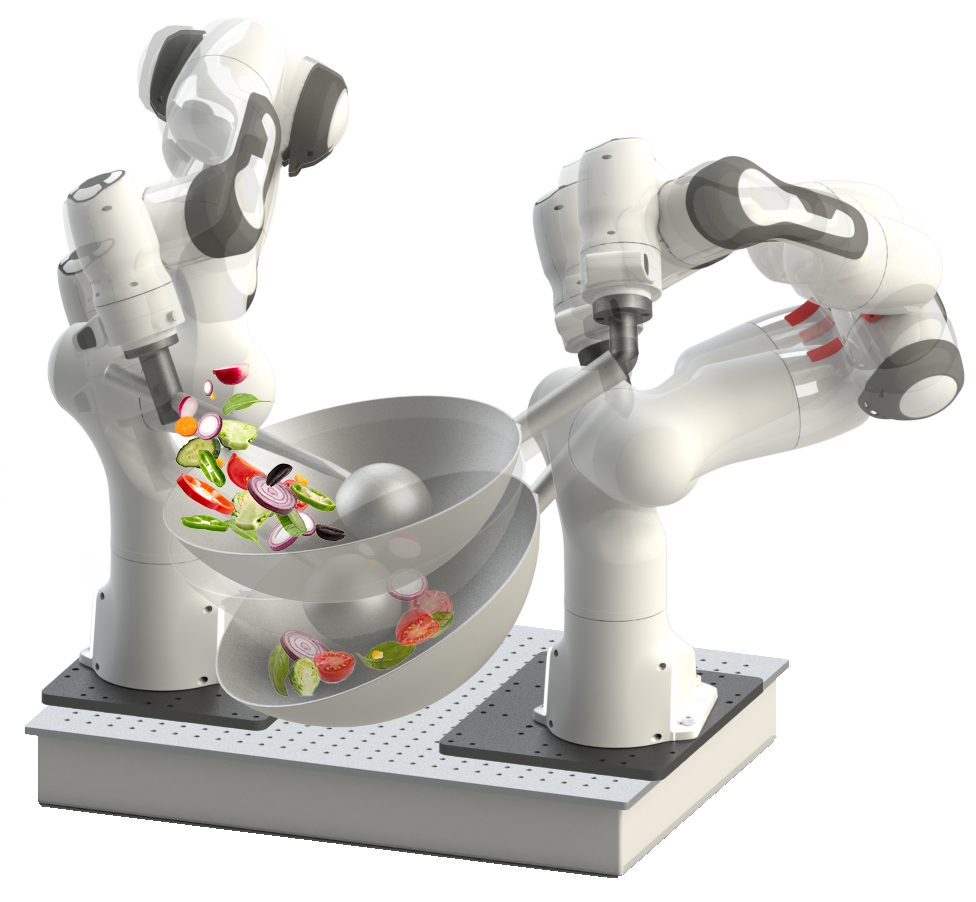}
	\caption{Robot stir-fry is a non-prehensile manipulation of semi-fluid objects which requires highly dynamic movements and continuous bimanual coordination in a long time series.}  
	\label{robotstirfry}
\end{figure}

Domestic service robots have been developed considerably in recent years \cite{fukuda2011advanced}, while the creation of a robot chef in the semi-structured kitchen environment remains a grand challenge. Food preparing and cooking are two important activities that take place in the household, and a robot chef who can follow arbitrary recipes and cook automatically would not only be practical but also bring new interactive entertainment experience. Several works propose to use a bimanual robot to realize food preparing processes, such as peeling motions \cite{dong2021food} or serving traditional Swiss raclette \cite{pignat2019bayesian}. Recently, a project called RoDyMan \cite{ruggiero2018nonprehensile} aims at a non-prehensile dynamic manipulation of deformable objects, mainly focusing on pizza making via a bimanual robot. In this letter, we are concerned about solving the problem of human-like robot cooking with the stir-fry movement, which is a significant and complicated bimanual skill in Chinese cooking art. It requires the dual arms of the chef or robot to bring a wok and a spatula, respectively, as shown in Fig. \ref{robotstirfry}. And it aims to realize a desired visual and cooking effect by rolling and tossing the content objects through the coordination of this two cookware.

Programming robot skills manually is a laborious effort and can only achieve one specific movement at a time. The most popular way is to learn from human demonstration (LfD). It aims to learn an optimal robot control policy that can generate trajectory following the distribution of demonstrations \cite{schaal1999imitation}\cite{argall2009survey}. There are already some skill learning solutions for single robot arm, such as Dynamic Movement Primitives (DMPs) \cite{ijspeert2013dynamical}, Probabilistic Movement Primitives (ProMPs) \cite{paraschos2013probabilistic} and Task-parameterized Gaussian Mixture Model (TP-GMM) \cite{calinon2016tutorial}. While in a bimanual setting, the coordination between dual arms becomes the core problem of the manipulation task. The existing bimanual coordination solutions are mostly built on kinematics with relevant coordinate frames. For example, the Compliant Movement Primitives in a bimanual setting can only solve symmetric control problem. It requires a complete task description, which can be separated into absolute and relative movements \cite{batinica2017compliant}. 

However, when handling the deformable objects, there does not exist a fixed relationship between dual arms, and the system is not possible to model as relative movements. What makes the task much trickier is that the contents in the wok comprise a mixture of liquid and solid, which we define as the deformable \textit{semi-fluid} content in this letter. The manipulation of it is even more difficult than ordinary deformable objects, especially the precise estimation of its state. Therefore, we pay more attention to achieving the manipulation through the coordination of dual arms and propose a proper control approach for these kinds of tasks. Unlike previous approaches, we regard robot bimanual coordination as a sequence transduction problem. By doing this, it is simplified as a single-arm control problem with a coordination module, and the proposed method can build on the advantages of existing single-arm skill learning methods. 

Sequence transduction problems like Neural Machine Translation (NMT) \cite{koehn_neural_2017} and Text-to-speech (TTS) \cite{tan_survey_2021} have been well studied in the Natural Language Processing (NLP) field. Their solutions from Wavenet-based \cite{oord_wavenet_2016} to Transformer-based \cite{vaswani_attention_2017} are all designed to capture the temporal correlation of the sequences. However, in bimanual coordination tasks, the spatial correlation between poses of dual arms also needs to be considered, which is used to be achieved by graph neural networks (GNNs) recently \cite{wu2020comprehensive}. Thus, it is reasonable to combine graph structures with the Transformer for capturing bimanual spatio-temporal relationship. There are already some existing works about combining graph and transformer, like \cite{yun_graph_2020}\cite{cai_graph_2020}, but they are proposed either for a graph sequence or for where the input sequence is represented as a graph, none of them can deal with the relationship between input and output sequences, and there are no relative applications in robotics.

Thus, by combining the neural network (NN) based coordination module and single-arm skill learning method, we propose an approach for manipulating the semi-fluid contents by learning bimanual stir-fry skill from demonstrations. The detailed contributions are as follows:

\begin{itemize}
\item To achieve an adjustable deformable object manipulation, we decouple dual arms into different roles: a leader and a follower. The leading movement is adjusted by DMP, and the corresponding following movement is generated via a proposed coordination module.
\item The coordination module --- \textit{Structured-Transformer}, which comprises a Graph and Transformer network, continuously couples the dual arms by capturing the spatio-temporal relationship between demonstrated dual-arm movements.
\item Our framework can realize robot stir-fry, a novel bimanual coordinated non-prehensile manipulation task of semi-fluid objects, and the adjustable capability is verified by the simulator and real robot system.
\end{itemize}



\section{Task Definition}
\begin{figure}[htp]
	\centering
	\includegraphics[width=1\linewidth]{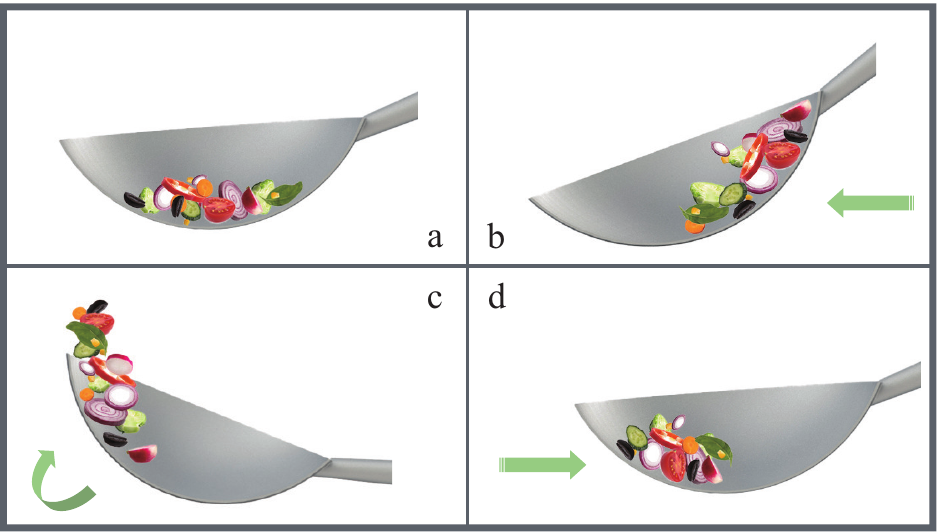}
	\caption{Starting with the static phase $a$, the single cycle of the rhythmic stir-fry movement is separated by three different phases, which are pushing forward $b$, rotating $c$ and pulling back $d$.}  
	\label{pendulm}
\end{figure}
In this work, we investigate the semi-fluid object manipulation by a bimanual robotic system \textit{without} explicit dynamical models of the task. The dual arms interact indirectly via a wok (left arm) and a spatula (right arm), and both of them shape the deformation of the semi-fluid object together. To make the task clearer, we describe the trajectory of the stir-fry movement and its difficulties, and finally define the objective.

\subsection{The general trajectory of stir-fry movement}\label{IIA}
	By observing plenty of teaching videos of human chefs, we summarize the canonical pattern of this movement and separate the trajectory of the wok into three different phases shown in Fig \ref{pendulm}, which are pushing forward $b$, rotating $c$, and pulling back $d$. Starting with the static phase $a$, these motions are rhythmic in a way of $b\to c\to d\to b$. On the other hand, the trajectory of the spatula is more unconstrained, it allows a relatively high range of movement between demonstrations. Its general trajectory is constantly switching between contact with the wok and free motion and affects the deformation by its agitation relative to the wok. There is no doubt that the rhythmic left-arm movement is the main factor of the relative movement of the semi-fluid content inside the wok, but the assistance of spatula also makes contributions to deformation and is an integral part of this bimanual skill. As long as the relative displacement of the semi-fluid content is similar to the one manipulated by a human in each specific phase, then we can regard it as a successful stir-fry.

\subsection{Problems of realizing stir-fry movement by robot}
	Achieving stir-fry with a robot is an intractable task, since many related control problems are still not fully developed. First, stir-fry is a \textit{non-prehensile} manipulation where the state of the operated content is subject only to unilateral constraints and the dynamics of both the food content and the end-effector, as well as the related kinematics \cite{ruggiero2018nonprehensile}. Second, the content in the wok are various in size, weight and stickiness. Due to the difficulty of modeling, the manipulation consequences are hard to obtain. Thus, it is difficult to design a general control method. Third, since dual arms \textit{interact indirectly} via cookware, the modeling of the robot system is heavy manual work and inconvenient to migrate to other cookware. Finally, the movement of stir-fry is \textit{highly dynamic} and requires \textit{continuous bimanual coordination} throughout the task.

\subsection{Objective}
\begin{figure*}[ht]
	\centerline{\includegraphics[width=1.00\linewidth]{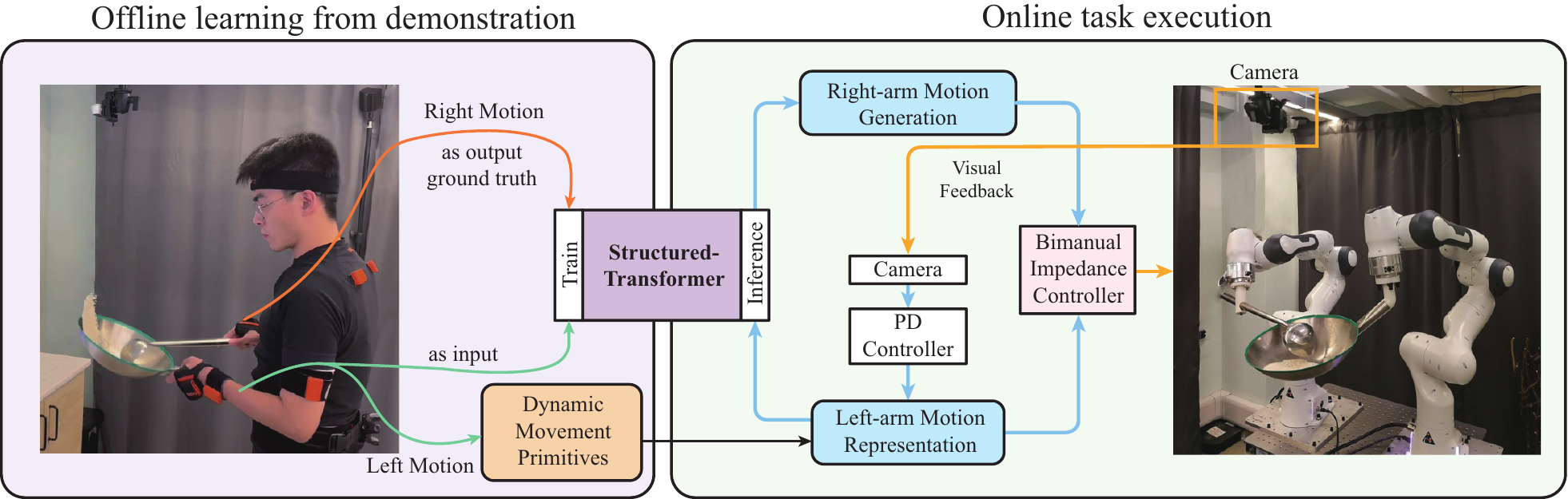}}
	\caption{The proposed decoupled coordination framework comprises two parts: offline learning from demonstration and online task execution. In the offline process, dual arms are decoupled into different roles and treated by a classical method and a neural network method separately. Here we adopt a DMP and propose a spatio-temporal neural network --- Structured-Transformer for learning the coordination. In the online process, left-arm motion is adjusted according to the visual feedback and the corresponding right-arm motion is generated by the pre-trained Structured-Transformer model based on the left-arm motion.}
	\label{Decoupled_Framework}
\end{figure*}
In this letter, we focus on learning the bimanual robot manipulation of deformable objects from human demonstrations, in particular, learning the robot stir-fry with semi-fluid content. Limited by the techniques of precisely estimating the task-specified semi-fluid content, we simplify this procedure and do not put lots of effort into it. The influence of being hard to model deformable objects is compensated for by proposing an adjustable and coordinated control framework. Therefore, dual arms can adjust their movements until the content reach a desired state according to the visual feedback.

\section{Bimanual Coordination Mechanism of Non-prehensile Manipulation}
\subsection{Bimanual non-prehensile manipulation of deformable objects}
In typical manipulations, like grasping tasks \cite{li2017reinforcement}\cite{chen2018dexterous}, there is no obvious relative motion between objects and robots. The infinitesimal motions of the object are restricted by the end-effector, through either form closure or force closure. While tasks like stir-fry, clothes folding and pushing only involve unilateral constraints, and the state changing relies on the dynamics of both the object and the robot, as well as their related kinematics and the (quasi-)static forces \cite{ruggiero2018nonprehensile}. Non-prehensile manipulation is more suitable for an unstructured anthropic environment, especially in domestic service. Although the demand for service robots growing fast, the solutions of non-prehensile manipulation tasks remain underdeveloped. Existing works aim to solve this by setting non-prehensile manipulation primitives and separating the task into subtasks \cite{lynch2003control}. However, this approach is only for specific tasks, rather than a general framework. Ruggiero \textit{et al.} share a similar idea and propose a high-level planning architecture to do the task decomposition automatically. They achieve an autonomous pizza-making task with a RoDyMan platform \cite{ruggiero2018nonprehensile}. Though they have used a humanoid robot, there is few complicated bimanual coordination and high dynamic motion in making pizza, which is both necessary in tasks like stir-fry. 

To achieve stir-fry with the robot, it is crucial that the robot can figure out the causal relationship between its motions and the deformation of objects, as well as the way of bimanual coordination. A proper visual feedback system like RoDyMan can satisfy the former purpose, while it is still challenging to learn a bimanual non-prehensile manipulation in which the dynamics of both arms are essential to the object deformation. Despite reinforcement learning methods \cite{mnih2015human}\cite{lillicrap2015continuous} are popular and can learn robot control policies from scratch, they all heavily rely on simulators with excellent physical simulation performance which is intractable when dealing with semi-fluid objects. A large sim2real gap will cause the policy learned in the simulator to be meaningless in the physical world \cite{peng2018sim}. Besides, since we cannot give an explicit definition of this kind of complicated movement, it is hard to analyze them from a theoretical aspect and difficult to program them manually. To solve this, we need to combine \textit{pure learning methods} and \textit{human prior knowledge} and try to let the robot learn its unique movement representation and coordination from simple demonstrations, rather than learning them from scratch or pre-defining them. Therefore, we plan to guide the robot's movement primitives learning through human demonstrations, and propose a bimanual coordination framework for learning adjustable deformable objects manipulation. 

\subsection{Different roles of dual arms in stir-fry}
\begin{figure*}[ht]
	\centerline{\includegraphics[width=0.80\linewidth]{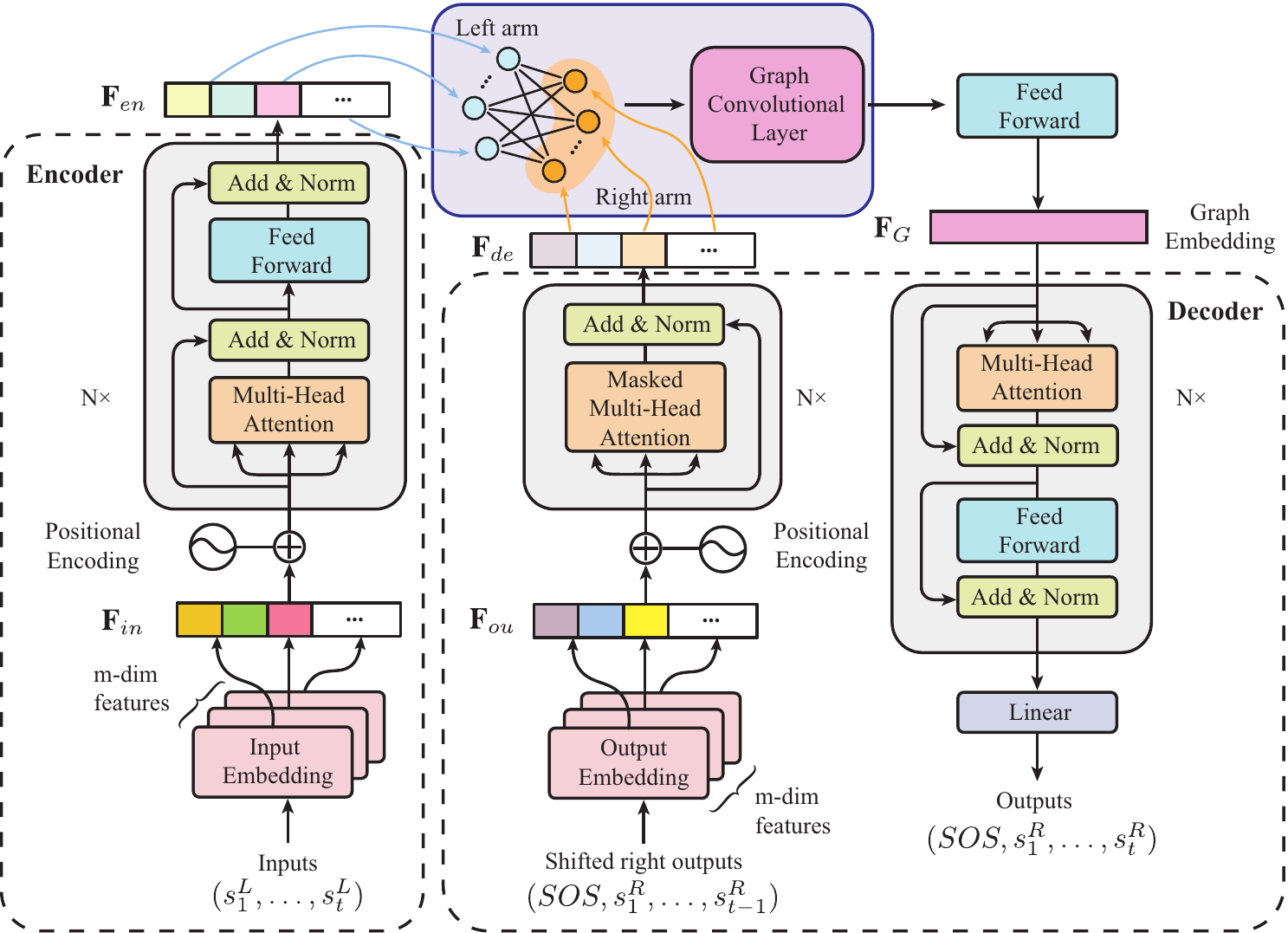}}
	\caption{Structured-Transformer embeds a Graph Convolutional Layer in the halfway of the Transformer decoder to consider the robot hardware structure explicitly. The input and output features can extend to $m$ dimensions, which can include force and visual information. Here, we use the 6-dim pose as input and output features. The left part of the decoder embeds the shifted right ground truth of the output, which is a training technique in recurrent neural networks called teacher forcing.}
	\label{Structured_Transformer}
\end{figure*}
As mentioned in Sec. \ref{IIA}, the left-arm movement has a fixed rhythm and it is hardly influenced by the outside environment. While the right-arm movement highly depends on the left arm. It is natural to set the dual arms as different roles: the left arm as a leader, and the right arm as a follower. Therefore, we propose a decoupled bimanual coordination framework, as illustrated in Fig. \ref{Decoupled_Framework}. 

The learning processes are starting from several human stir-fry demonstrations. We firstly separate the data into left and right motions. The primitive of left-arm motion is learned by the classical Movement Primitives (MP) method to get a robust performance. Here we adopt DMP, which allows the robot to perform the same behavior in different start or/and stop poses. 
Unlike the left arm, the motion of the right arm is generated based on the left. 
Their motions are learned in different ways, and rollout in succession: adjust the left motion via DMP first, and then generate the corresponding right motion based on it. To guarantee a proper continuous coordination, we propose a Graph and Transformer based neural network model --- \textit{Structured-Transformer}. It can capture the spatio-temporal relationship between the dual-arm sequential motions and the robot bimanual structure. By giving the robot this ability, it can perform the same bimanual behavior without spatial and temporal restriction.

To sum up, we have used three methods to solve the problem of stir-fry. First, the idea of LfD is used to solve the difficulty of reproducing the highly dynamic stir-fry motion. Second, we decouple this task and represent the left arm motion via DMP for robustness and adaptation. Finally, we propose a learning model for general continuous coordination. 
\subsection{Representing demonstrations as primitives}
As mentioned above, the motion of left arm is represented by DMP for generating adjustable movements which follows the demonstrated behavior. The DMP starts with a simple dynamical system, and is transformed into a weakly nonlinear system by a learnable autonomous forcing term. The formula is defined as follows:
\begin{equation} \label{cond1}
	\begin{aligned}
		&\tau^2\ddot{\bm{y}} = \alpha_y(\beta_y(\bm{g}-\bm{y})-\tau\dot{\bm{y}})+ \bm{f} \\
        &\bm{f}(x, \bm{g})  = \frac{\sum^{N}_{i=1}{\psi}_i\cdot\omega_i}{\sum^{N}_{i=1}{\psi}_i }x(\bm{g}-\bm{y_0})
	\end{aligned}
\end{equation}
where $\dot{x}=-\alpha_x x$ is an introduced canonical dynamical system. ${\psi}_i=\exp{-h_i(x-c_i)^2}$ defines a Gaussian basis function centered at $c_i$, where $h_i$ is the variance. $\alpha_y, \beta_y, \alpha_x$ are gain terms. $\bm{y}, \bm{y_0}, \bm{g}$ refer to the current pose, the initial pose and the target pose in the Cartesian space, respectively. These variables construct the nonlinear force term $\bm{f}$, which is the crux of the DMP method and makes the dynamical system follow some desired trajectories. The $(\bm{g}-\bm{y_0})$ and $\tau$ terms show the temporal and spatial generalization capabilities.

In order to make the stir-fry process adjustable, the DMP we adopt is a discrete version with Gaussian basis functions, rather than using the rhythmic DMP. Thus, in each cycle, motion of the wok (left arm) is separated into three parts as the $b, c, d$ phases in Fig. \ref{pendulm}, and represented by three different DMPs. By tuning the poses of connection points between each phase, the left-arm motion can be carried out at different speeds and forces, but with the same behavior as the demonstrations.

\subsection{Learning continuous coordination by combining Graph and Transformer}
Although the dual-arm movements can be represented individually by using MP-based methods, the performance of coordination is not guaranteed, especially in tasks like stir-fry which have indirect interactions between dual arms. Therefore, it is necessary to propose a specific module for learning the inherent relationship between them. This module should generalize from demonstrations to motions after DMP adjustment. The network structure is illustrated in Fig. \ref{Structured_Transformer}.

Since we have given the arms different roles in manipulation, the problem of coordination can be abstracted into a sequence transduction problem between the movements of both arms. Suppose the motion sequence of the left arm is known to be $\mathbf{s}^L=\{s_0^L, s_1^L, \dots, s_T^L\}$, where $\{s_t^L, t\in 0\sim T\}$ refers to the 6-dim pose of left arm at time $t$. We wish to obtain the corresponding right motion sequence by learning this conditional formula:
\begin{equation} \label{cond2}
	\begin{aligned}
		p(\mathbf{s}^R)=\prod_{t=1}^T p(s^R_t|s_1^L, \dots, s_t^L, s_1^R, \dots, s_{t-1}^R)
	\end{aligned}
\end{equation}

It means that the generated pose of the right arm at time $t$ is related not only to the current left-arm pose but the historical poses of dual arms. In \textit{Structured-Transformer}, the combination of $s_t^L$ and the historical left movements is encoded by the \textit{Encoder}, while the historical right trajectory $(s_1^R, \dots, s_{t-1}^R)$ is shifted right with a zero value start-of-sequence ($SOS$) and inputs to the \textit{Decoder}. The final output of this model is $(SOS, s_1^R, \dots, s_{t}^R)$. Since its structure is derived from Transformer, we only describe the modifications relative to the original model.

\textbf{\textit{Encoder}:} The principal part of the Transformer encoder remains unchanged, which is composed of a stack of $N = 3$ identical layers. In order to deal with multidimensional features in motion data simultaneously, we use several input embedding layers to represent the 6-dimensional features of the current left-arm state separately, where $s^L_t = (x_t^L, y_t^L, z_t^L, \gamma_t^L, \alpha_t^L, \beta_t^L)$, $(x_t^L, y_t^L, z_t^L)$ for position and $(\gamma_t^L, \alpha_t^L, \beta_t^L)$ for (\textit{roll, pitch, yaw}). These input embeddings are then concatenated together in order as a single tensor $\textbf{F}_{in}\in \mathbb{R}_{b\times sl \times h_{em}}$ before the positional encoding process, where $b$ for batch size, $sl$ for sequence length, $h_{em}$ for the dimension of the embedding layer. After the process of encoder, a corresponding tensor $\textbf{F}_{en}\in \mathbb{R}_{b\times h_{en} \times h_{em}}$ is obtained, where $h_{en}$ for the dimension of the encoding layer.

\textbf{\textit{Decoder}:} The decoder of the vanilla Transformer is split into two parts after the decoder masked self-attention. 
After $N$ layers of this masked attention, the shifted right historical right trajectory is decoded into the tensor $\textbf{F}_{de}\in \mathbb{R}_{b\times h_{de} \times h_{em}}$. The main modification of the decoder occurs in the input of the encoder-decoder module. In vanilla Transformer, the key $K$ and value $V$ is the encoder tensor $\textbf{F}_{en}$, and the query $Q$ is the decoder tensor $\textbf{F}_{de}$. However, in \textit{Structured-Transformer}, we adopt the graph embedding $\textbf{F}_{G}$ as its $K, V, Q$. After a linear layer, the predicted next pose of the right arm is obtained. It is worth mentioning that since each pair of encoder and decoder tensors corresponds to the same graph, the training process still can be executed in a parallel way. 

\textbf{\textit{Graph}:} We adopt a graph structure to consider the spatial relationship between bimanual movements explicitly. Both outputs from the encoder and left decoder ($\textbf{F}_{en}, \textbf{F}_{de}$) are separated into corresponding features so that we can use them as node feature to construct a graph. The edges between nodes refer to the effect from one of the left feature to one of the right, such as how the position change of the left arm in $z$-axis influence the change of the right arm in $\gamma$. Only one dimension of the graph construction in Fig. \ref{Structured_Transformer} is given as an example. This graph represents the whole robot state of that current frame. The graph embedding process is designed to get a vector representation $\textbf{F}_{G}$ of the whole graph via Graph Convolutional Layer and feed-forward network. 

Overall, motions in this model are transformed from sequence vector to sequence graph and then back to the sequence vector. Thus, formula \ref{cond2} is modified into a structured version:

\begin{equation}
	\begin{aligned}
		p(\mathbf{s}^R)=\prod_{t=1}^T p(s^R_t|G_1, \dots, G_{t-1}, s_t^L)
	\end{aligned}
\end{equation}
where $G_t=(N_t, E_t)$, $N_t = (g(s_t^L), g(s_t^R))$, $g(\cdot)$ refers to transformation before the Graph module.

\section{Experiments}
\subsection{Setup}
The experiments contain several parts: human demonstration data collection, bimanual coordination network training (Sec. \ref{VII}), simulation (Sec. \ref{VIII}), and real robot stir-fry (Sec. \ref{VIV}). Due to the high cost and safety issue of robot experiments, it is vital to first verify motions in the simulation, and then deploy them on the real robots. Thus, we built a simulation platform via PyBullet to simulate the real robot platform, which contains dual Franka Panda robotic arms. This platform not only applies to this task but is also useful for studying other kitchen skills. In this section, we describe some preparatory works: the process of demonstration data collection and the visual feedback system.

\textbf{\textit{Data Collection}:} Due to the requirement of high dynamic and proper coordination of bimanual stir-fry, the demonstration data of this movement was recorded by an Xsens suit and its supporting software. The motion-capture data we adopt is from two of the IMU sensors, which are mounted on the left and right hands (see the offline learning process in Fig. \ref{Decoupled_Framework}). These data are transformed into a rosbag format for further network learning. During the data collection process, an RGB-D camera is attached to the ceiling to record the temporal states of the semi-fluid contents in the wok. These video data are used to analyze the proper deformation and displacement of the semi-fluid contents for evaluation.

\begin{figure*}[ht]
	\centerline{\includegraphics[width=0.85\linewidth]{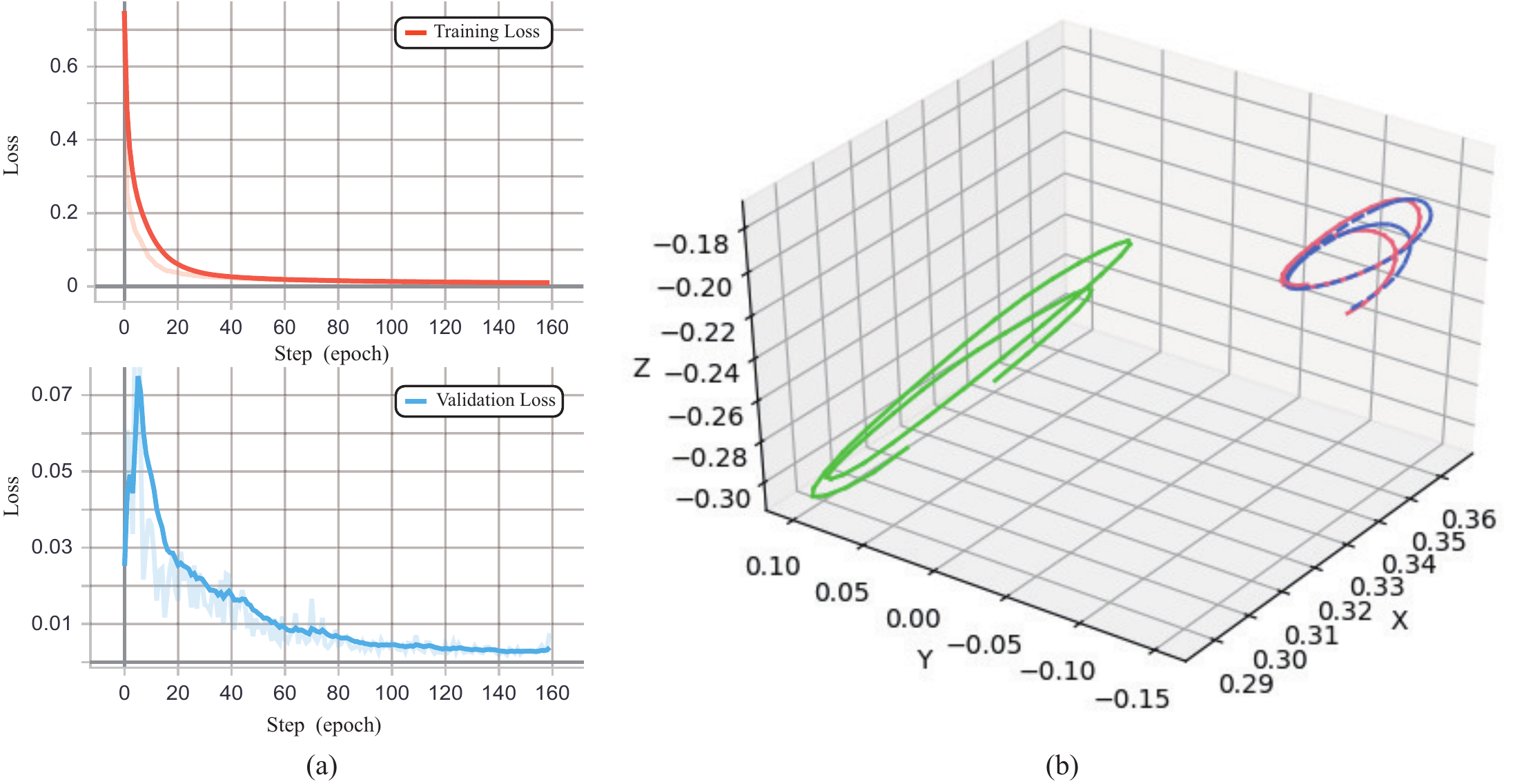}}
	\caption{The performance of the coordination learning method. (a) shows the loss curves of both training and validation processes, the solid lines are the results after smoothing with a factor of 0.6, and the light lines refer to the original values; (b) shows a period of rollout by Structured-Transformer, the green and red lines represent the ground truth of left-arm and right-arm movements, while the blue line is a fully generated right-arm motion according to the left.}
	\label{learning_performance}
\end{figure*}

\textbf{\textit{Visual Feedback}:}
To evaluate the performance of our motion and adjust the trajectory, a visual feedback is required. As shown in the online execution process of Fig. \ref{Decoupled_Framework}, a Realsense camera is fixed to the top to get the top-down map of the wok when conducting the real robot experiment. The images are captured at a frequency of 10Hz, and the area of semi-fluid content is segmented by image processing techniques (see Fig. \ref{visual}). The wok is first located by the green edge line and Hough circle transform. Then the content is segmented by the Watershed algorithm. The relative displacement is represented by the distance between the center of segmented area and the wok. The video of demonstration is also processed in the same way, and the average value of the maximum relative displacement in each cycle is used as the target value. 

\subsection{Learning performance}\label{VII}
\textbf{\textit{Implementation details}:} The parameters of \textit{Structured-Transformer} are listed as follow: hidden dimension $h_{en}=h_{em}=h_{de} = 210$, stack layer $N=3$, feature dimension $m=6$ and 3 attention heads. We adopt dynamic time warping (DTW) between the generated right movement poses and the demonstration as the loss function and train the network via backpropagation with the Adam optimizer, linear warm-up phase for the first 5 epoch, a decaying learning rate afterward, and a dropout value of 0.1. The Graph module uses a 1-layer Graph Convolutional Network and a feed-forward network for getting an embedding of the whole graph. The normalization of the network input influences its performance, as also maintained in \cite{graves2013generating}\cite{zhang2019sr}. So we normalize the motions by subtracting the mean and dividing by the standard deviation of the train set. 

During the training process, we use Teacher Forcing technique \cite{mihaylova2019scheduled} for a paralleled training. It uses the ground truth from a previous time step as the decoder input, rather than the previous output of the network. Since the coordination learning network aims to obtain a generalized relationship between the wok and spatula in stir-fry, we need to use a different demonstration as the validation set. The validation is executed after every epoch of training. While different from the Teacher Forcing used in the training process, the decoder in the validation process is auto-regressive, which uses the prediction in time step $t-1$ as the decoder input of time step $t$ and does not rely on the ground truth. Therefore, the correlation between the loss curve of training and validation can reflect the generalization of the model in different motions of the same behavior. We use scaled data enhancement in the training set, where the scale is [0.5 $\sim$ 1.5] and the interval is 0.1. The validation set uses data with scales of 0.65 and 0.75. From Fig. \ref{learning_performance} (a), we can find that the model has a good generalization to the movement within the training data distribution. The final movement conditional generation performance is demonstrated by the rollout of the testing set with a scale of 0.85, which is shown in Fig. \ref{learning_performance} (b). The fully generated right-arm movement is smooth and close to the ground truth, and the normalized DTW value is 0.003. Since both the left-arm and right-arm movements are not existing in the training set and validation set, we can regard it as a good coordination generalization, which attributes to the learning of dual-arm inherent relationship.
\begin{figure*}[ht]
	\centerline{\includegraphics[width=0.95\linewidth]{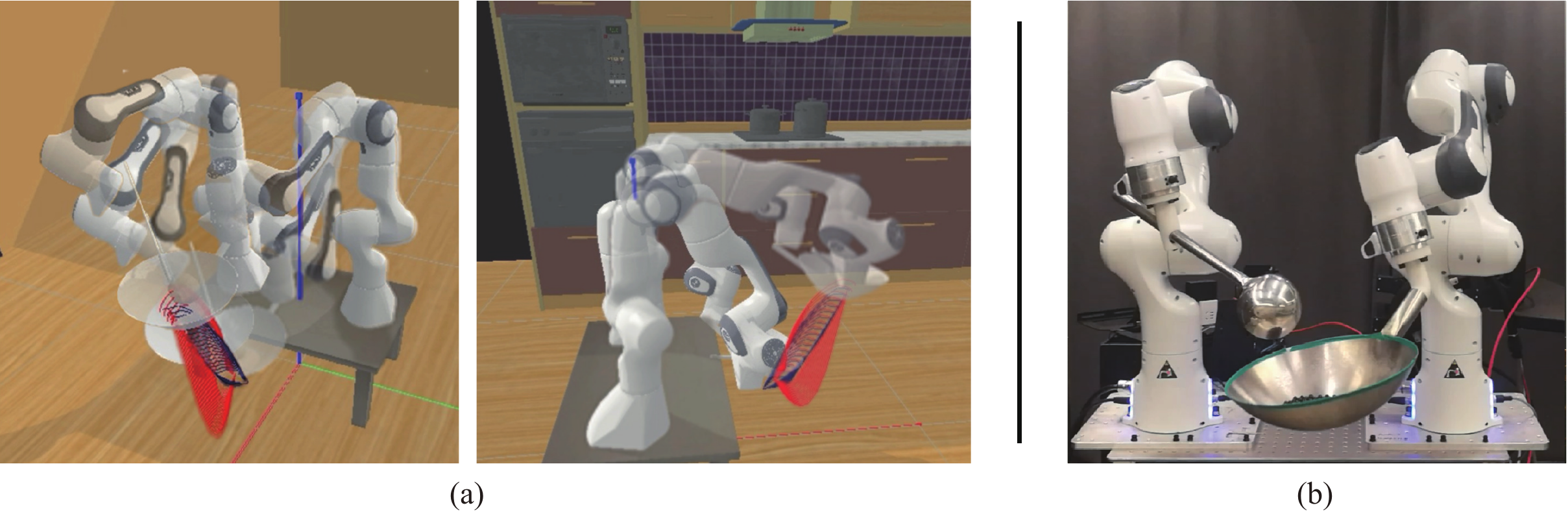}}
	\caption{(a) shows a simulation experiment of dynamical adjustment from two different perspectives, the black curves refer to different left-arm motions which are adjusted through DMP, and the red curves are the corresponding right-arm motions generated by the \textit{Structured-Transformer}; (b) shows the experiment on the real bimanual robot platform.}
	\label{Simulation}
\end{figure*}
\noindent
\subsection{Simulation}\label{VIII}
\begin{figure}[ht]
	\centerline{\includegraphics[width=0.95\linewidth]{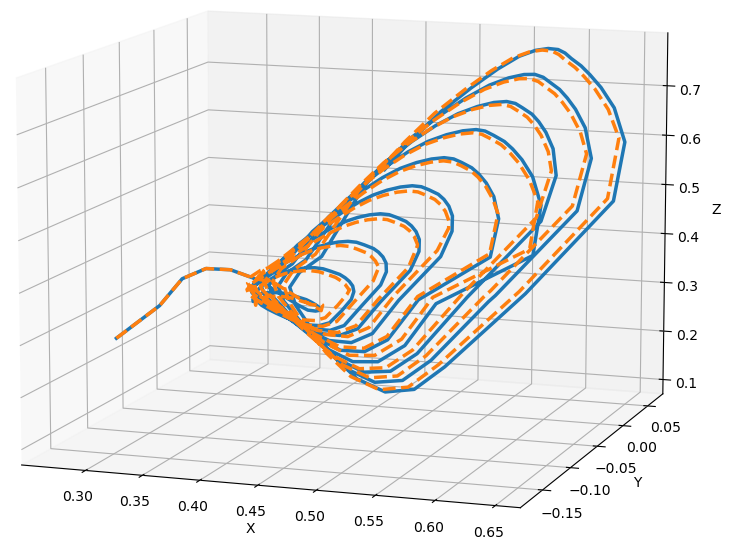}}
	\caption{The coordination performance can be illustrated by comparing the desired and the real right-arm motion in the simulator. The blue line refers to the desired movements, while the line in orange is the real right-arm movements after coordinated with the left.}
	\label{sim_compare}
\end{figure}
A dual-arm robot platform with a wok and spatula is built in the PyBullet simulator. Two Franka Emika Panda robots are fixed at a table in the same position as the real-world setup, which enables us to verify the safety of the motion generated from our method. To make the scene more realistic, we also build a virtual kitchen environment, and it allows us to study more kitchen skills in the future.

We first test a single left-arm adaptive wok tossing using DMP and then combine the DMP with \textit{Structured-Transformer} to generate a less collision bimanual manipulation after adjusting the left-arm movements. The coordination simulation is illustrated in Fig. \ref{Simulation} (a) from two different perspectives. Black curves show the dynamically adjusted left-arm movements, the connection points of three phases in each cycle are increasing progressively. Red curves refer to the corresponding right-arm movements generated by the \textit{Structured-Transformer}. Their coordinates are based on the end center of each cookware. Since the Pybullet has a simulation of collision, we can figure out the performance of coordination by comparing the right-arm movements between the desired and the real one, as shown in Fig. \ref{sim_compare}. We use the same evaluation index as the learning performance to assess the extent of collision and get a normalized DTW value of 0.0049. 

\begin{figure}[ht]
	\centerline{\includegraphics[width=1\linewidth]{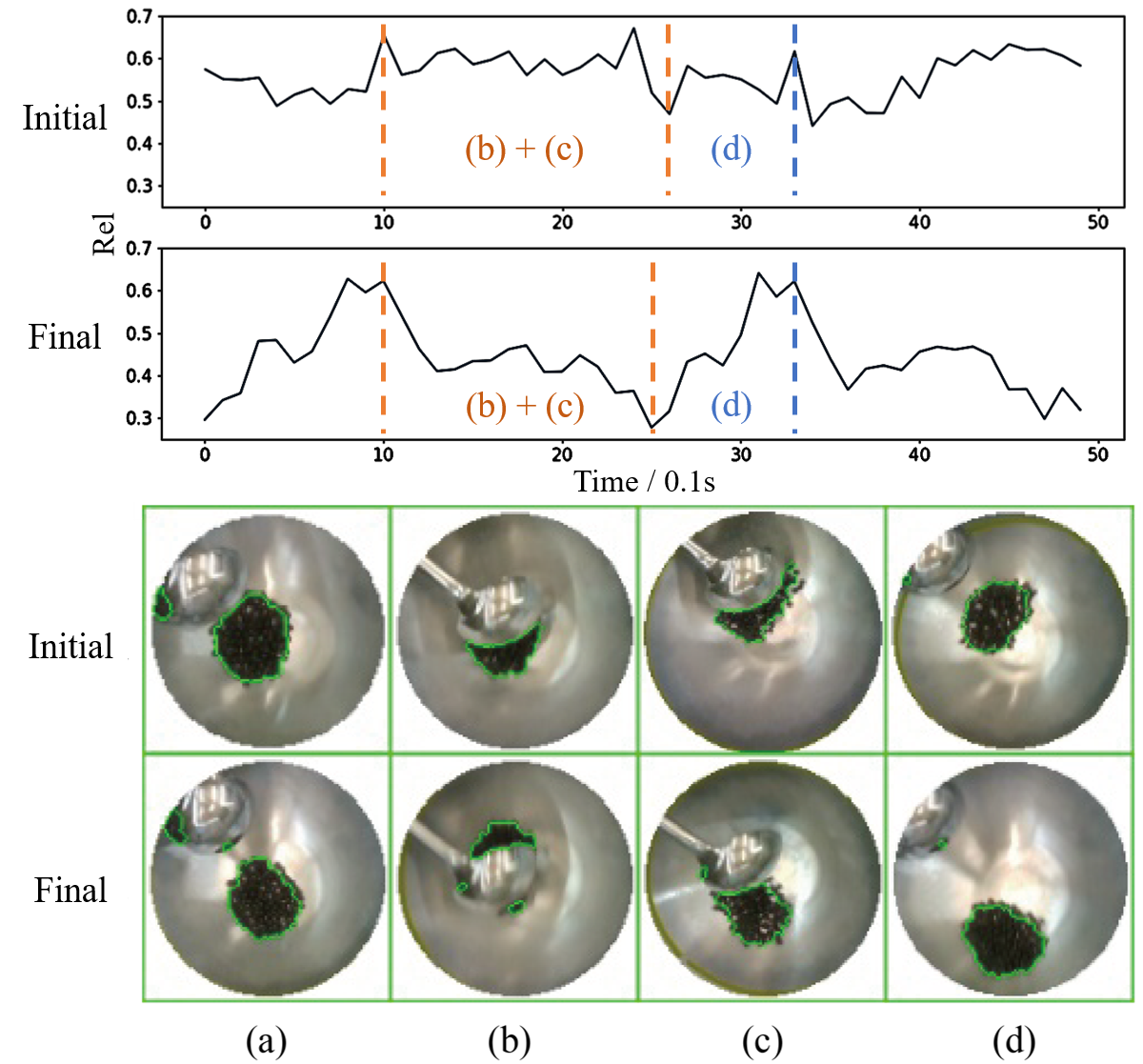}}
	\caption{The area of semi-fluid contents is segmented by the image processing techniques, and the relative displacement is represented by the distance between the center of the segmented area and the wok.}
	\label{visual}
\end{figure}
\subsection{Real robot stir-fry}\label{VIV}
The real robot platform contains two Franka Emika Panda robots, a 500g wok, a spatula and their connectors, and also the visual feedback device, as shown in Fig. \ref{Decoupled_Framework} and Fig. \ref{Simulation} (b). We set the world coordinate  on the middle of the table surface. Two Franka Emika Panda robots are placed on along the Y-axis at an interval of 58 cm in the same posture. The RGB-D camera is fixed on at a coordinate with (50 cm, 0, 130 cm). 

Before executing the rollout movements, we measure a proper coordinate mapping between the IMUs attached on human hands and the end-effectors. Then the rollout movements are transformed into the robot coordinate. The execution of bimanual movements is based on an impedance controller. Besides, the DMP adjustment is conducted offline by a PD controller. Thus, the real robot experiment can be regarded as the repeat of following steps: 

\begin{enumerate}
	\item [(1)] Execute the generated dual-arm movements by the bimanual robot system;
	\item [(2)] Get the visual feedback during the execution, evaluate the relative displacement of the semi-fluid contents;
	\item [(3)] Obtain a new left-arm movement through DMP;
	\item [(4)] Generate a new right-arm movement according to the adjusted left-arm movement via the pre-trained \textit{Structured-Transformer} model.
\end{enumerate}

The visual feedback and the relative displacement from real robot stir-fry are shown in Fig. \ref{visual}. The y-axis of the first two rows is the relative displacement according to the wok, this is a rate value between 0 and 1. The smaller this value is, the semi-fluid content is closer to the front edge of the wok. The four columns of the last two rows correspond to the four phases mentioned in Fig. \ref{pendulm}. The final relative displacement curve and the segmentation show that after adjusting through the proposed method, the state of the content finally meets the requirements of a proper stir-fry movement.



\section{Conclusion}
In this letter, we define a novel non-prehensile deformable object manipulation task which perform stir-fry using a bimanual robot. We propose an approach to solve the task in a decoupled manner by regarding the dual arms as different roles and learning successively. Then, a learning model is proposed to obtain the inherent spatio-temporal relationship between the wok and spatula. This model combines the benefit of Graph structure and Transformer sequence learning, and uses them to represent the bimanual robot structure and the temporal information in the motion. The simulation and real robot platform we built will be a foundation of future research in kitchen skill learning tasks like stir-fry. The way we decouple the bimanual motion and the proposed relational coordination learning method may give a new inspiration to other bimanual manipulation tasks, as well as human-robot collaborative manipulation tasks.

However, we only consider poses of demonstrations, while contact forces also exist in the stir-fry. Thus, higher dimensional information will be introduced to learn a more humanoid motion in kitchen skills, such as visual, myoelectric signals. Besides, as mentioned in this letter, the estimation of semi-fluid contents is simplified as two-dimensional image segmentation, and we only use the relative displacement as the desired target. Another lateral camera needs to be introduced for checking whether the content will leak.

\bibliographystyle{ieeetr}
\bibliography{output}

\end{document}